%% file: 0acl2020.tex
\documentclass[11pt,a4paper]{article}
\usepackage[hyperref]{acl2020}
\usepackage{times}
\usepackage{latexsym}

\usepackage{array}
\usepackage{float}
\usepackage{graphicx}
\usepackage{caption}
\usepackage{enumitem}
\newcommand{\PreserveBackslash}[1]{\let\temp=\\#1\let\\=\temp}
\newcolumntype{C}[1]{>{\PreserveBackslash\centering}p{#1}}
\newcolumntype{R}[1]{>{\PreserveBackslash\raggedleft}p{#1}}
\newcolumntype{L}[1]{>{\PreserveBackslash\raggedright}p{#1}}

\usepackage{microtype}

\aclfinalcopy % Uncomment this line for the final submission
\def\aclpaperid{145} %  Enter the acl Paper ID here

\title{ \textit{``where is this relationship going?''}: Understanding Relationship Trajectories in Narrative Text}

\author{Keen You \and Dan Goldwasser \\
    Department of Computer Science, Purdue University \\
    \texttt{\{you54, dgoldwas\}@purdue.edu} \\}

\date{}

\begin{document}
\maketitle
\begin{abstract}
We examine a new commonsense reasoning task: given a narrative describing a social interaction that centers on two protagonists, systems make inferences about the underlying relationship trajectory. Specifically, we propose two evaluation tasks: Relationship Outlook Prediction MCQ and Resolution Prediction MCQ. In Relationship Outlook Prediction, a system maps an interaction to a relationship outlook that captures how the interaction is expected to change the relationship. In Resolution Prediction, a system attributes a given relationship outlook to a particular resolution that explains the outcome. These two tasks parallel two real-life questions that people frequently ponder upon as they navigate different social situations: \textit{``where is this relationship going?''} and \textit{``how did we end up here?''}. To facilitate the investigation of human social relationships through these two tasks, we construct a new dataset, \textbf{\textit{Social Narrative Tree}}, which consists of 1250 stories documenting a variety of daily social interactions. The narratives encode a multitude of social elements that interweave to give rise to rich commonsense knowledge of how relationships evolve with respect to social interactions. We establish baseline performances using language models and the accuracies are significantly lower than human performance. The results demonstrate that models need to look beyond syntactic and semantic signals to comprehend complex human relationships.
\end{abstract}

\input{1Intro}
\input{2RelatedWork}
\input{3Dataset}

\input{4TaskFormulation}
\input{5Experiment}
\input{6ErrorAnalyses}
\input{7Summary}

\bibliography{anthology,acl2020}
\bibliographystyle{acl_natbib}

\end{document}

%% file: 1Intro.tex
\section{Introduction}

% task motivation
A relationship between two people is constantly being shaped by their social interactions \cite{duck1994meaningful}. 
For example, if two people have a proper conversation after a heated argument, they may become more intimate as they understand each other better. On the other hand, they may insist on their own views and part ways. For humans, being able to reason about relationship trajectories is crucial in achieving personal goals while avoiding conflicts with other people. This skill is formally termed as \textit{social competence}, which humans easily acquire \cite{rubin1995competence}.

% why is understanding relationships difficult and how this task is different from existing tasks
In contrast to humans' innate ability to perceive social situations, machines struggle in developing social understanding. For example, machines' performance in reasoning about motivation and emotional reactions given a short context is significantly lower compared to human performance \cite{sap2019socialIQa}. Moreover, identifying the intents and reactions of characters in narratives poses another challenge for machines~\cite{goyal2010automatically,chaturvedi2016ask,rahimtoroghi2017modelling,RocstoryPsychology}. Reasoning about relationship trajectories adds yet another layer of difficulty as in addition to the challenge of identifying the implicit factors that guide characters' behavior, such as intents, reactions and mental states, systems also need to understand how these elements collectively contribute to the evolution of relationships. This additional layer of understanding is challenging primarily because the impact of a particular event on a relationship is unique to each social scenario. For instance, an argument can help resolve differences or deepen them, depending on the personalities, intents and reactions of the characters involved in it.

% an argument can make or break a relationship depending on how it is resolved, which in turn depends of a number of other factors including intents, reactions, personalities, current relationship and so on.

% why current datasets are lacking/not suitable for relationships
The resources and methods discussed in current work do not directly address these challenges. %Event2Mind \cite{rashkin2018event2mind} provides a knowledge base mapping between activities and their intents and outcomes, which 
A fixed set of mappings from events to relationship impact, in a style similar to Event2Mind \cite{rashkin2018event2mind}, is not sufficient to capture the uniqueness of each social situation.
%For this reason, having a fixed set of mappings from events to relationship impact, similar to Event2Mind \cite{rashkin2018event2mind} that maps phrase to intents and reactions, will not be sufficient to capture the uniqueness of such impact in each situation. 
Similarly, SocialIQA \cite{sap2019socialIQa} inspects various elements individually but does not unify them into a single force that influences a relationship. Other current datasets \cite{rocstories,swag,HellaSwag} cover a broad spectrum of commonsense knowledge, but do not focus on the social aspect. Although it can be annotated for such inspection such as in the investigation of intents and reactions \cite{RocstoryPsychology}, this method is not pertaining to relationship analyses because a meaningful change may not always be present. 
%[DG]: Add back if there is space
For example, here is a description of a social scenario sampled from the ROCStories corpus: \textit{Tina decided on going hiking with her friend Tony. They both decided on a difficult path. Upon ascension, Tina fell and cut her leg. They both decided it was too dangerous for them to continue. Tony carried Tina to the car and decided on mini-golf instead.} The story provides a few facts regarding a social interaction but does not provide sufficient details to draw conclusions about how the relationship between the two protagonists is changed because of the interaction. Unanswered questions include \textit{``did Tina appreciate Tony's gesture?''} and \textit{``did they bond as a result?''}. 
All these deficiencies in using current resources and methods to study human relationships call for the need of new resources that encode implicit relationship trajectories that are driven by the interweaving of social elements. 

% why is our dataset capable
In this work, we introduce \textbf{\textit{Social Narrative Tree}}\footnote[1]{The dataset is publicly available at \url{https://github.com/karenacorn99/Social-Narrative-Tree}}, a corpus of 1250 social narratives documenting a variety of social interactions, each centers on two protagonists. It is built incrementally from ten seeds in five narrative stages and the story branches into five different paths at each stage. This effectively captures the different possibilities which a relationship trajectory can take on at each diverging point. We choose the narrative stages -- seed, buildup, climax, resolution and outlook - based on previous narrative analyses \cite{freytag1896freytag,prince1973grammar,Labov1997SomeFS}, ensuring that intensity varies at different points of the narrative which simulates fluctuating intensity levels in real-life social situations and thus provides a natural space for relationships to develop in. Using \textit{\textbf{Social Narrative Tree}}, we set up two evaluation tasks, Relationship Outlook Prediction MCQ and Resolution Prediction MCQ. In each task, the text corresponding to Relationship Outlook or Resolution is removed from the stories for prediction based on other stages. The branching of each story at different stages creates a natural notion of similarity among stories, which can be measured by the number of stages two stories share. This facilitates the setup of confounding choices as the wrong candidates are not completely irrelevant in terms of textual information. This forces systems to truly understand the underlying relationship trajectories instead of relying solely on language modelling. The best accuracy achieved by our BERT-based model~\cite{devlin2019bert} (\textasciitilde60\%) is significantly lower than human performance (\textasciitilde80\%). These experimental results demonstrate that models need to look beyond syntactic and semantic signals to comprehend complex human relationships.

In summary, our contributions are as follows: (1) we introduce \textit{\textbf{Social Narrative Tree}}, a corpus of social narratives with contextualized social elements contributing to relationship trajectories.
(2) we introduce new evaluation frameworks using Relationship Outlook Prediction and Resolution Prediction, with meaningful confounding candidate choices that force systems not to depend only on textual information.
(3) we establish baseline performances using language models and justify the importance of looking beyond textual information in understanding human relationships.

%% file: 2RelatedWork.tex
\section{Related Work}

Manually constructed scripts are used to represent structured knowledge in 1970-80s \cite{ScriptSchankAbelson1977}. Subsequently, narrative event chains, unsupervised generation of such representations are introduced. Narrative event chains are sequences of events revolving around one central protagonist \cite{unsupervised_learning_of_narrative}. Stemming from narrative event chains is the narrative cloze task where one event chain is removed and systems are required to fill in the blanks \cite{unsupervised_learning_of_narrative}. The task is further refined to multiple choice form - Multiple Choice Narrative Cloze (MCNC) where different choices are randomly sampled from events that do not belong to the chain \cite{WhatHappensNext}. Swaf Af \cite{swag} and Hella Swag \cite{HellaSwag} are datasets for next event prediction, containing multiple choice questions covering a wide range of grounded situations that are constructed from video captions. Furthermore, \cite{rocstories} create ROCStories, a corpus of 50k commonsense stories each of five-sentence long and propose Story Cloze Test where systems are required to select the most plausible ending for an incomplete narrative. In our work, Relationship Outlook Prediction and Resolution Prediction are multiple-choice tasks of similar motivation with a focus on social implications.

Various social elements are extensively studied in many works. \cite{Atomic} present ATOMIC, a knowledge graph of commonsense knowledge with 877k descriptions in free text form and focuses on 'if-then' relationships of causes, effects and attributes. Automatic construction of such knowledge base is explored in Comet~\cite{Bosselut2019COMETCT}. With ATOMIC as the foundation, SocialIQA, a dataset of 38,000 multiple-choice questions about social situations is constructed \cite{sap2019socialIQa}. Event2Mind is a corpus of phrasal verbs that is constructed to support the examination of the intents and reactions of common situations \cite{rashkin2018event2mind}. \cite{VAD} presents NRC VAD Lexicon, a corpus of 20,000 English words with human ratings of valence, arousal and dominance while NRC Affect Intensity Lexicon (AIL) provides emotional categories and associated real values for approximately 6,000 English words \cite{WordAffectIntensities}. SocialSent \cite{SocialSent} assigns tokens sentiment scores with contexts taken into consideration. One set of lexicons is constructed from the subreddit r/relationships which we use in our error analyses. Other works model relationships, between literary characters~\cite{iyyer2016feuding} or countries~\cite{han2019permanent}, described in text using an unsupervised neural model, by mapping them to a latent space.
%between nations from news articles and point out that verbal predicates encode valuable information about relationships.
Elements of psychology in ROCstories are analyzed in \cite{RocstoryPsychology} where ROCStories are annotated with motivations and emotional reactions of the characters involved. We create \textit{\textbf{Social Narrative Tree}} with various social and psychological elements embedded in each story and these elements together contribute to the rise and fall of a relationship.

%% file: 3Dataset.tex
\section{Collaborative Construction of Social Narratives}
% \subsection{Social Narrative Structure}  
%\subsubsection{Narrative Stages}
Our main contribution in this paper is the construction of a social narrative corpus consisting of short stories each describing the evolution of a relationship between two characters. The corpus is designed to capture how different social behaviors result in different relationship outcomes. To accomplish this goal we follow a fixed narrative structure consisting of five stages: seed (\textit{exposition}), buildup (\textit{rising action}), climax, resolution and outlook (\textit{denouement}). The stages are based on ``Freytag's Pyramid''~\cite{freytag1896freytag} and other more recent work analyzing repeating narrative structures~ \cite{prince1973grammar,Labov1997SomeFS}. 
Fundamentally, a minimal story consists of an initial state, a final state and an intermediate state that transitions the initial state to the final state \cite{prince1973grammar}. More specifically, the intermediate state can be further broken down into individual stages to make a story more appealing and informative. Particularly, we adopt the phases of orientation clause, Most Reportable Event~(MRE) and resolution \cite{Labov1997SomeFS} to make up the intermediate state, corresponding to the buildup, climax and resolution stages in our framework. %Overall, the initial relationship state is introduced in seed and the events that happen in buildup, climax and resolution shapes the relationship into its final state as describe in the outlook. 
The relationship in each story develops as the intensity of the narrative fluctuates. 

The dataset is constructed in a collaborative way to help capture the impact of different behaviors at each narrative stage. We use Mechanical Turk, and provide annotators with a partial story created independently, and ask them to complete the next stage. Each MTurker is limited to provide three responses in total at each stage with all three responses belonging to different prompts.

\subsection{Crowdsourcing Framework}
\textbf{Seed.}~ The seed is a one-sentence description that introduces two protagonists, their initial relationship and a general social scenario that they are going to be involved in. The dataset is built on 10 initial seeds, which are created from the first sentences of 10 randomly selected stories from the ROCStories dataset~\cite{rocstories} by adding names and removing collective nouns. We inspect and reselect the randomly sampled seeds to ensure that there are no repetitions in social scenarios. This helps to elicit a variety of social behaviors and associated relationship impact. Among the 10 pairs of protagonists, seven pairs have the same gender, two pairs have different genders and one pair includes the name ``\textit{Sam}'' which MTurkers have different gender interpretations. All the seeds have the same initial relationship description, \textit{``friend''}, but the social scenario that the two protagonists are involved in provides additional information on the intimacy level between the two people (at one's home versus a public place). Each seed contains a predicate connecting the two protagonists. Among these predicates, \textit{``asked''} and \textit{``invited''} are active, \textit{``receive''} is passive while \textit{``realized''}, \textit{``were''} and \textit{``went to''} have a neutral voice. \ 

\noindent\textbf{Buildup.}
%Buildup is the first stage that is built using crowd-sourcing. 
We instruct MTurkers to provide a one-sentence continuation of the seed, which provides further information on the relationship between the protagonists or on the social scenario. Five different responses are collected for each seed. We specifically state in the instruction that only information relevant to the relationship of the protagonists should be given. This ensures that the plot is compact and the development of the relationship is the driving force of the narrative. We first programmatically filter responses that are completely irrelevant or do not meet the length requirement. We then manually filter all the remaining responses and re-annotate those that do not meet the requirements until the desired number of buildups are collected for each seed. We allow the addition of characters other than the two main protagonists in the responses, diversifying the types of events present in the dataset. 

\noindent\textbf{Climax.}
%Climax is the next stage built using crowd-sourcing. We create five partial narratives for each seed by combining each seed with every one of its buildups, resulting in 50 partial narratives in total.
We create 50 partial narratives, consisting of the 10 seeds and their associated buildups, which serve as the input for the collection of climax events. %The climax stage correspond to the Most Reportable Event for the story~\cite{MRE}. 
We prompt MTurkers to provide a two-sentence continuation for each seed and buildup combination as the climax event, which we define as an event that has a significant impact on the relationship between the people involved. Five responses meeting the requirement are collected for each partial narrative, forming 250 seed-buildup-climax combinations.  

\noindent\textbf{Resolution and Outlook.}  
We prompt MTurkers to provide a two-sentence continuation of the partial story as the resolution and a one-sentence continuation of the resolution as the relationship outlook. The purpose of resolution is to resolve, either successfully or not, the conflict brought up in the partial narratives. Relationship outlook states the effect of the complete interaction described in the narrative on the relationship between protagonists. Five resolution-outlook pairs are collected for each one of the 250 partial narratives, capturing how different choices of resolving conflicts can impact relationships. At the end of this stage, 1250 complete social narratives are created.

\subsection{Dataset Overview}

We illustrate the overall tree structure of Social Narrative Tree in Fig. 1 and present some basic statistics of the dataset in Tab. 1. In Tab. 2, we summarize the most frequent five predicates for each narrative stage. In this analysis, we use NLTK \cite{BirdKleinLoper09} to extract and lemmatize verbs and discard a list of stop words including \textit{be}, \textit{will}, \textit{do} and \textit{can}. The most used verbs in each stage are consistent with the general purpose of each stage. The seed initiates a social interaction \textit{(invite, ask, receive)} and in buildups, the desires \textit{(want, ask, decide)} and mental states \textit{(think, feel)} of the protagonists are revealed. Climax is where the actual relationship-changing event happens \textit{(start, go, get)} while resolution and outlook contain a mixture of further actions \textit{(say, get, make)}, mental states \textit{(feel)} and a notion of change \textit{(decide, become, realize)}. 

\begin{figure*}[t]
\centering
\includegraphics[scale = 0.15]{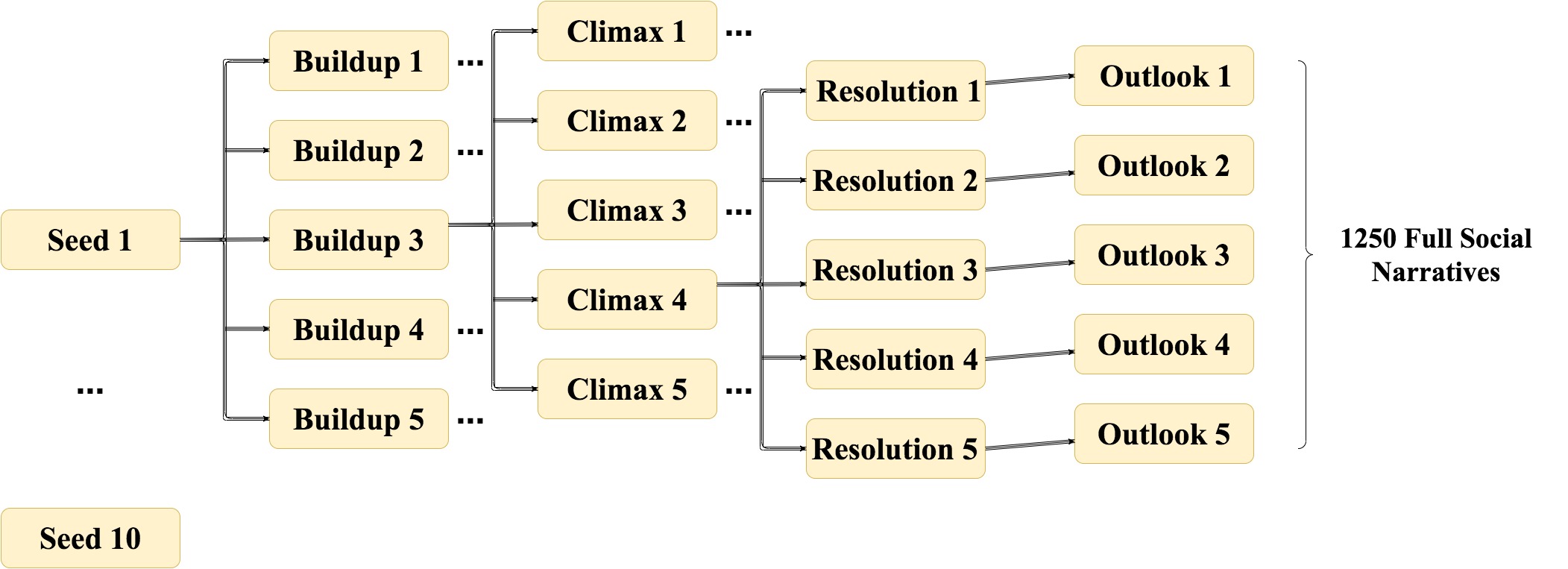}
\caption{Visualization of Social Narrative Tree's overall structure.}
\end{figure*}

\begin{table*}[t]
\centering
\resizebox{125mm}{!}{%
{\small
\begin{tabular}{|L{2.0cm}|C{1.5cm}|C{1.5cm}|C{1.5cm}|C{1.6cm}|C{1.5cm}|C{1.5cm}|} \hline
                    & \textbf{Seed} & \textbf{Buildup} & \textbf{Climax} & \textbf{Resolution} & \textbf{Outlook} & \textbf{Full Story} \\ \hline
\# sentences        & 1    & 1       & 2      & 2          & 1       & 7          \\ \hline
\# uni. instances & 10   & 50      & 250    & 1250       & 1238    & 1250       \\ \hline
\# vocab              & 70   & 427     & 1583   & 3366       & 2294    & 4474       \\ \hline
avg token \#    & 12.5 & 24.7    & 35.2 & 30.7     & 19.1    & 122.2    \\ \hline
max token \#        & 15   & 40      & 92     & 108        & 57      & 211        \\ \hline
min token \#       & 10   & 14      & 12     & 6          & 3       & 66        \\ \hline
\end{tabular}%
}}
\caption{\label{tab:table-name}Social Narrative Tree basic information.}
\end{table*}

\begin{table}[h]
\centering
\resizebox{70mm}{!}{%
{\small
\begin{tabular}{|l|l|l|l|l|}
\hline
\textbf{seed}                                                                             & \textbf{buildup}                                                                       & \textbf{climax}                                                                  & \textbf{resolution}                                                                 & \textbf{outlook}                                                                            \\ \hline
\begin{tabular}[c]{@{}l@{}}invite\\ ask\\ go \\ receive \\ realize\end{tabular} & \begin{tabular}[c]{@{}l@{}}want\\ think \\ ask \\ decide \\ feel\end{tabular} & \begin{tabular}[c]{@{}l@{}}start\\ go \\ get \\ want\\ say\end{tabular} & \begin{tabular}[c]{@{}l@{}}decide\\ feel \\ tell \\ say\\ get\end{tabular} & \begin{tabular}[c]{@{}l@{}}decide\\ become \\ feel \\ realize \\ make\end{tabular} \\ \hline
\end{tabular}}}
\caption{Most frequent five verbs for each stage.}
\end{table}

\begin{table}[h]
\centering
\resizebox{75mm}{!}{%
{\small
\begin{tabular}{|l|l|l|l|l|l|}
\hline
         & \textbf{seed} & \textbf{buildup} & \textbf{climax} & \textbf{resolution} & \textbf{outlook} \\ \hline
positive & 0    & 32      & 115    & 678        & 650     \\ \hline
negative & 0    & 13      & 106    & 377        & 269     \\ \hline
neutral  & 10   & 5       & 29     & 195        & 331     \\ \hline
\end{tabular}}}
\caption{Sentiment polarity distribution at each narrative stage.}
\end{table}

\subsection{Relationship Trajectories in \textbf{\textit{Social Narrative Tree}}}
We carry out various exploratory analyses on \textbf{\textit{Social Narrative Tree}} to examine the relationship trends present in the dataset with respect to social elements. The results validate that the stories contain a broad range of possible ways that relationships can unfold under different circumstances. In addition, the relationship trajectories present in the stories also exhibit some associations with different social elements, verifying that valuable social behavior commonsense knowledge is embedded in the narratives.

% a lot of different trends, not just good good good
\noindent\textbf{Sentiment Polarity.} We associate a sentiment polarity label with each stage using TextBlob \cite{loria2014textblob}.
%We use TextBlob to give each stage for every story a sentiment polarity label, particularly, we assign a positive label to a positive score, a negative label to a negative score and a neutral label to a score of 0.
The sentiment polarity distribution for each stage is shown in Tab. 3. The sentiment polarity distribution for climax is fairly equal (115 pos vs 106 neg) but positive resolutions and outlooks significantly outnumber negative resolutions and outlooks (678 pos vs 377 neg, 650 pos vs 269 neg). This indicates that in \textbf{\textit{Social Narrative Tree}}, social interactions are more likely to project positively. We further inspect overall relationship trends by breaking down stories into different combinations of polarities across stages and present the results in Tab. 4 and 5. From Tab. 4, we see that both positive and negative buildup branch equally into positive and negative climax (350 pos vs 350 neg, 150 pos vs 145 neg), giving both trends opportunities to develop and reducing the chance of monotonous plots. From resolution to outlook in Tab. 5, however, both positive and negative resolution are more likely to lead to positive outcomes. This observation could potentially tie back to people's general goal of avoiding conflicts with other people~\cite{rubin1995competence} which MTurkers instill into the stories.

%This potentially relates to the general social observation that at an early stage of a social interaction, the general polarity can change more freely. It contrasts with results in Tab.4, where a positive resolution is much more likely to lead to a positive outlook than a negative outlook, possibly implying that at later stages of a social interaction, the general polarity is subject to more specific social norms. Interestingly, negative resolution leads to more positive outlook than negative outlooks, possibly suggesting humans' nature to avoid conflicts in social situations.

\begin{table}[h]
\parbox{.45\linewidth}{
\centering
{\small
\begin{tabular}{|l|l|l|}
\hline
\textbf{buildup} & \textbf{climax} & count \\ \hline
pos     & pos    & 350   \\ \hline
pos     & neg    & 350   \\ \hline
neg     & pos    & 150   \\ \hline
neg     & neg    & 145   \\ \hline
\end{tabular}}
\caption{Sentiment polarity trends between buildup and climax.}
}
\hfill
\parbox{.50\linewidth}{
\centering
{\small
\begin{tabular}{|l|l|l|}
\hline
\textbf{res} & \textbf{outlook} & count \\ \hline
pos        & pos     & 403   \\ \hline
pos        & neg     & 123   \\ \hline
neg        & pos     & 163   \\ \hline
neg        & neg     & 105   \\ \hline
\end{tabular}}
\caption{Sentiment polarity trends between resolution and outlook.}
}
\end{table}

%\begin{table}[H]
%\centering
%\begin{tabular}{|l|l|l|l|l|}
%\hline
%climax & resolution & outlook & count      \\ \hline
%pos    & pos        & pos     & 226        \\ \hline
%pos    & neg        & pos     & 69         \\ \hline
%pos    & pos        & neg     & 50         \\ \hline
%pos    & neg        & neg     & 35         \\ \hline
%neg    & pos        & pos     & 142        \\ \hline
%neg    & neg        & pos     & 82         \\ \hline
%neg    & pos        & neg     & 55         \\ \hline
%neg    & neg        & neg     & 61         \\ \hline
%\end{tabular}
%\caption{Sentiment polarity distribution in different narrative stages.}
%\end{table}

% trajectory associated with social elements
\noindent{\textbf{Valence, Arousal, Dominance (VAD).}}
~For each stage we compute a vector of length three as the VAD representation \cite{VAD}. Within a stage, we retrieve the VAD scores for all tokens and take the maximum value for each dimension to represent the entire stage. We further associate VAD scores with general relationship trajectories represented using sentiment polarity by inspecting the VAD scores in different subsets of the corpus and present the results in Tab. 6 and 7. In Tab. 6, positive outlook is associated with higher valence and dominance in climax, resolution and outlook. It is also associated with lower arousal in climax and resolution. In Tab 7, positive climax is associated with lower arousal in climax but higher arousal in resolution. These results indicate that in \textbf{\textit{Social Narrative Tree}}, relationship trajectories are closely related to VAD which is an important social behavior indicator. This close relation can be transformed to rich commonsense knowledge if studied in-depth.

%\begin{table}[H]
%\centering
%%\begin{tabular}{|l|l|l|l|l|l|}
%\hline
%  & seed   & buildup & climax   & resolution & outlook  \\ \hline
%V & 0.9251 & 0.88344 & 0.875588 & 0.8732664  & 0.876808 \\ \hline
%A & 0.6408 & 0.72484 & 0.756044 & 0.7365392  & 0.653492 \\ \hline
%D & 0.6332 & 0.72458 & 0.714324 & 0.7245856  & 0.709952 \\ \hline
%\end{tabular}
%\caption{Average VAD for each stage.}
%\end{table}

\begin{table*}[t]
\centering
\resizebox{125mm}{!}{%
{\small
\begin{tabular}{|l|lll|lll|lll|}
\hline
                                                    \textbf{outlook}        &                             & climax                      &        &                             & resolution                  &        &                             & outlook &        \\ \cline{2-10} 
                                                            & \multicolumn{1}{l|}{V}      & \multicolumn{1}{l|}{A}      & D      & \multicolumn{1}{l|}{V}      & \multicolumn{1}{l|}{A}      & D      & \multicolumn{1}{l|}{V}      & \multicolumn{1}{l|}{A}       & D      \\ \hline
\begin{tabular}[c]{@{}l@{}}positive \end{tabular}  & \multicolumn{1}{l|}{0.8796} & \multicolumn{1}{l|}{0.7537} & 0.7189 & \multicolumn{1}{l|}{0.8900} & \multicolumn{1}{l|}{0.7372} & 0.7349 & \multicolumn{1}{l|}{0.9123} &  \multicolumn{1}{l|}{0.6721}  & 0.7409 \\ \hline
\begin{tabular}[c]{@{}l@{}}negative \end{tabular} & \multicolumn{1}{l|}{0.8697} & \multicolumn{1}{l|}{0.7627} & 0.7099 & \multicolumn{1}{l|}{0.8631} & \multicolumn{1}{l|}{0.7503} & 0.7195 & \multicolumn{1}{l|}{0.8395} & \multicolumn{1}{l|}{0.6656}  &  0.6797 \\ \hline
\end{tabular}}}
\caption{VAD scores of climax, resolution and outlook partitioned by outlook sentiment polarity.}
\end{table*}
\vspace{-20pt}
\begin{table*}[t]
\centering
{\small
\resizebox{125mm}{!}{%
 \begin{tabular}{|l|lll|lll|lll|}
\hline
                                                         \textbf{climax}  &                           & climax                      &        &                             & resolution                  &        &                             & outlook                     &        \\ \cline{2-10} 
                                                           & \multicolumn{1}{l|}{V}      & \multicolumn{1}{l|}{A}      & D      & \multicolumn{1}{l|}{V}      & \multicolumn{1}{l|}{A}      & D      & \multicolumn{1}{l|}{V}      & \multicolumn{1}{l|}{A}      & D      \\ \hline
\begin{tabular}[c]{@{}l@{}}positive \end{tabular}  & \multicolumn{1}{l|}{0.8973} & \multicolumn{1}{l|}{0.7612} & 0.7513 & \multicolumn{1}{l|}{0.8870} & \multicolumn{1}{l|}{0.7424} & 0.7292 & \multicolumn{1}{l|}{0.8827} & \multicolumn{1}{l|}{0.6559} & 0.7188 \\ \hline
\begin{tabular}[c]{@{}l@{}}negative \end{tabular} & \multicolumn{1}{l|}{0.8617} & \multicolumn{1}{l|}{0.7723} & 0.6911 & \multicolumn{1}{l|}{0.8630} & \multicolumn{1}{l|}{0.7253} & 0.7237 & \multicolumn{1}{l|}{0.8752} & \multicolumn{1}{l|}{0.6502} & 0.7034 \\ \hline
\end{tabular}}}
\caption{VAD scores of climax, resolution and outlook partitioned by climax sentiment polarity.}
\end{table*}

\vspace{16pt}
\noindent{\textbf{Affect Intensity.}}
Affect Intensity assigns a real value to a lexicon in one of the four dimensions -- joy, fear, anger and sadness \cite{WordAffectIntensities}. We assign a binary vector of length four for each stage for every story, indicating whether that particular dimension is present in the text for that stage. 
We analyze how different affect dimensions relate to general relationship trajectories by dividing the stories into two subsets using climax sentiment polarity. In each subset, we further separate the stories into two groups, one with positive trend and the other with negative trend -- positive trend means the story lands on a positive relationship outlook and vice versa. Within each group, we compute the percentage of resolutions that contain words from each Affect dimension. Fig. 2 and 3 display the results for the positive climax set and the negative climax set respectively. In Fig. 2,  resolutions in a positive trend contain a higher proportion of joy and a lower proportion of fear, anger and sadness. The distribution of Affect dimensions is similar in Fig. 3, except resolutions following a negative climax with a positive trend has a higher proportion of fear. Given a positive or negative relationship-changing event, this analysis relates mental states, another important social behavior indicator, in resolution to relationship trajectories after the event, which is an important aspect of social behavior knowledge.

\vspace{5pt}
\begin{figure}[H]
  \centering
  \begin{minipage}[b]{0.4\textwidth}
    \includegraphics[scale=0.37]{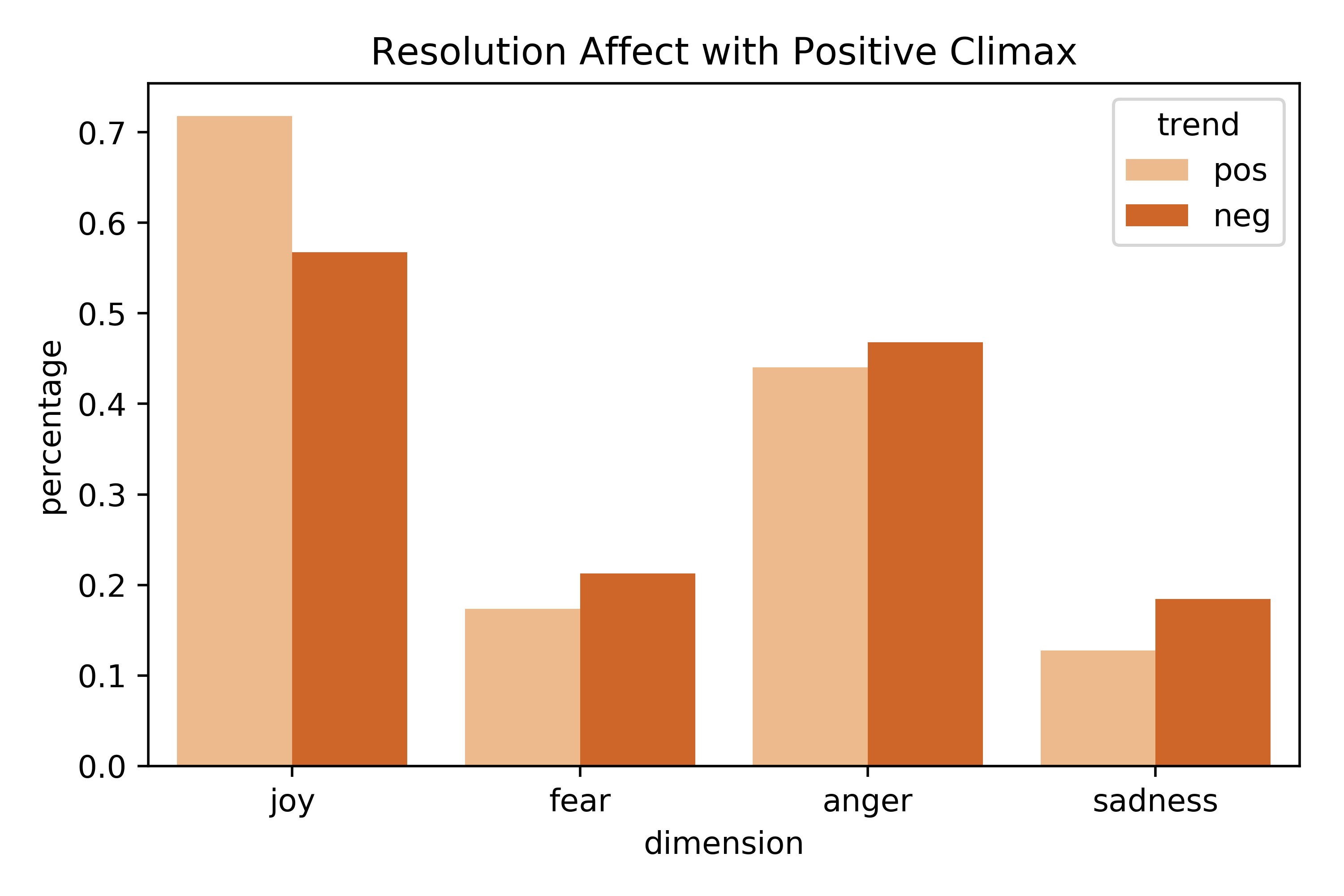}
    \vspace{-7pt}
    \caption{Resolution Affect dimensions in positive climax set.}
  \end{minipage}
  \hfill
  \begin{minipage}[b]{0.4\textwidth}
    \includegraphics[scale=0.37]{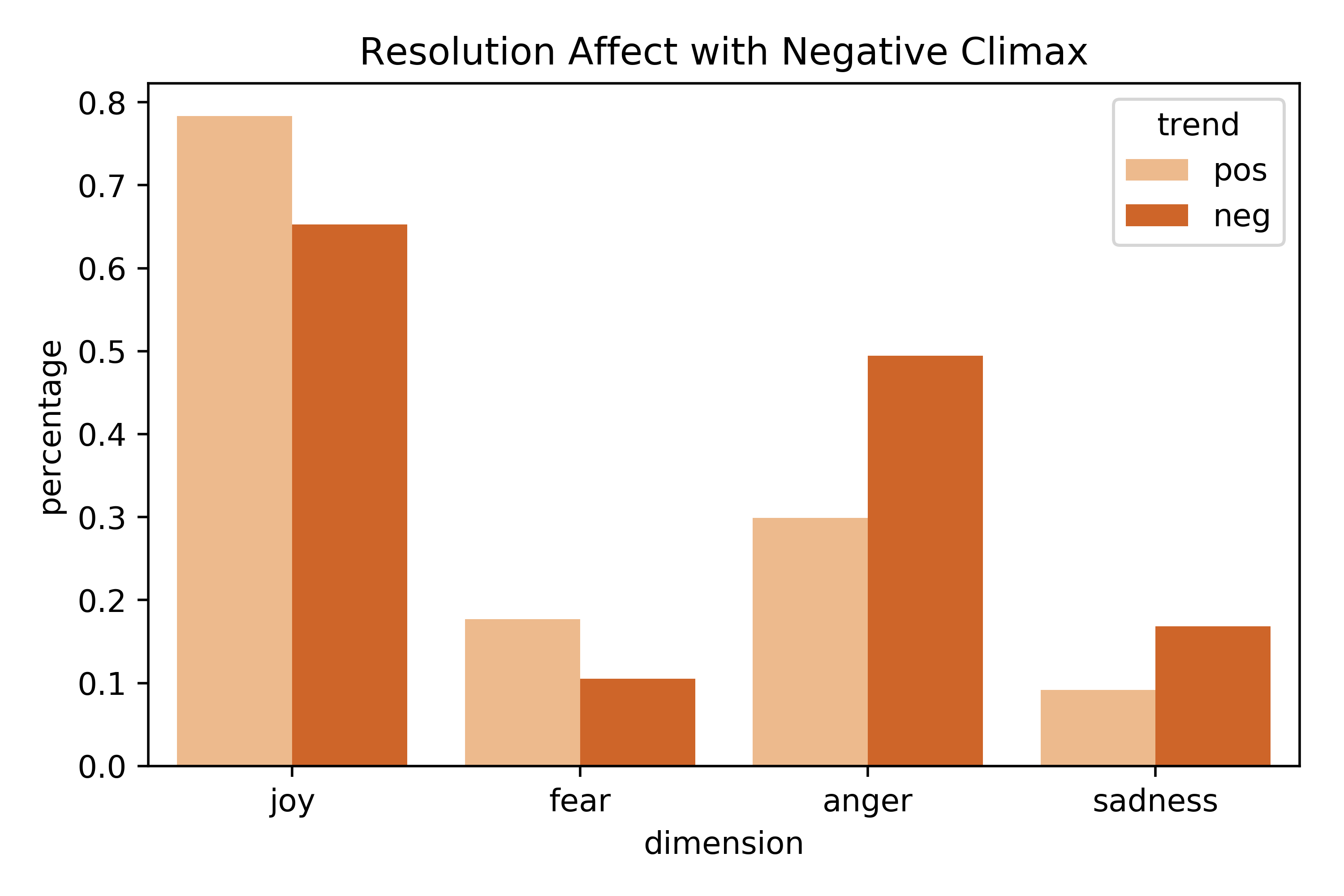}
    \vspace{-7pt}
    \caption{Resolution Affect dimensions in negative climax set.}
  \end{minipage}
\end{figure}

% condition on outlook polarity, how climax affect varies
%\begin{figure}[h!]
%    \centering
%    \includegraphics[scale = 0.6]{social relationship/images/OutlookPolarityClimax.png}
%    \caption{condition on outlook polarity, how climax affect varies}
%    \label{fig:my_label}
%\end{figure}

% start with good climax, how resolution affect varies with outlook polarity
%\begin{figure}[H]
%    \centering
%    \includegraphics[scale = 0.6]{social relationship/images/ResolutionAffectPosClimax.png}
%    \caption{}
%    \label{fig:my_label}
%\end{figure}

% start with bad climax, how resolution affect varies with outlook polarity
%\begin{figure}[H]
%    \centering
%    \includegraphics[scale = 0.6]{social relationship/images/ResolutionAffectNegClimax.png}
%    \caption{We take the subsets of stories with negative climax and does similar %computations as in the previous figure.}
%    \label{fig:my_label}
%\end{figure}

%% file: 4TaskFormulation.tex
\section{Task Formulation}

% 3.1 Predict outlook given resolution/seed+buildup+climax+resolution

%[DG] I think you are missing a discussion about the narrative structure that would give some context to the task definition. -- Why are the two stages interesting (outlook, resolution) what kind of common sense do they capture?
%[KY] \cite{Prince}, \cite{Labov} will explain further
%Why is it more than sentiment prediction? Similarity between sentences?  
% [KY] word2vec result analysis - similairty between sentences
\subsection{Relationship Outlook Prediction: \textit{``Where is this relationship going?''}}

In this task, Relationship Outlook is removed from stories and systems need to fill in the blank by choosing the most reasonable outlook from a pool of five candidates. Solving this task requires an understanding of social norms, particularly, what is expected of a relationship after a sequence of events happen in a particular social context. 
A sample question is shown below and choice D is the correct answer. %The starting relationship is "friend" and the general social situation is an ethnic celebration. The particular relationship-changing event is that James lied to Neil about the event attire and Neil was mad. The question that follows is how their friendship is changed because of this event. 
 In Choice A, the emotion of embarrassment is correctly inferred but ``felt more comfortable'' is inconsistent with the fact that Neil left the celebration. Choice B and Choice E are irrelevant to the given context. In choice C, ``not realize that the event was not a casual one'' is a correct statement with respect to the context but the expectation of them remaining friends is a socially incorrect deduction. 

\vspace{2pt}
\noindent \textit{Neil went to a celebration with his Vietnamese friend James. Neil was a little nervous as he had never been to an ethnic celebration before, however, James reassured him that it was going to be very casual. When Neil arrived, the event was anything but casual. As he looked around the room, he saw women in dresses and men in suits and he was wearing t-shirt and jeans! James insisted that what Neil was wearing was fine, but Neil still disagreed. Neil got angry at James for lying to him and then left the celebration in tears.} \newline
\newline
\textbf{\textit{What is the most likely relationship outlook?}}
\begin{enumerate}[label=(\Alph*),topsep=0.5pt,itemsep=-0.8ex,partopsep=1ex,parsep=1ex] 
\item \textit{Although at first Neil was embarrassed, throughout the night he felt more comfortable with James, and his confidence in his friend grew exponentially.}
\item \textit{James and Neil realized how much they had in common, and it strengthened their friendship.}
\item \textit{They remained friends because Neil had not realise that the event was not a casual one.}
\item \textbf{\textit{Neil felt like James could no longer be trusted after lying to him about the celebration, so he thought it would be best to stop talking to James.}}
\item \textit{They will remain friends and more than likely to place more bets on sports or challenges that come between each other, it's friendly competition of two friends' egos.}
\end{enumerate}

% 3.2 Predict resolution given outlook/seed+buildup+climax+outlook
\subsection{Resolution Prediction: \textit{``How did we end up here?''}}
After Resolution is removed from stories, systems need to reason about the cause of a relationship outlook by selecting one of the five resolution candidates. To select the correct resolution, systems need to understand the necessary conditions that will result in a particular relationship outcome.
%
%On a fine-grained level, there are two possible interpretations depending on the polarity of the final state. If the outcome is positive, systems need to have an understanding of what actions are required to achieve a relationship goal from a diverging point where previous stages have brought the protagonists to. If the outcome is negative, knowledge of cause and effects of social interactions is necessary in identifying the reason behind a negative consequence.
%
An example Resolution Prediction question is given below. Choice B is correct as it completes the narrative both logically and socially correct. Choices A and D are relevant, %in terms of her feelings are unknown to him 
but a romantic relationship would not have started if her feelings remain hidden. Choice C diverts from the context by mentioning a gift exchange whereas choice E is unlikely before the beginning of a romantic relationship.

\vspace{2pt}
 \noindent\textit{Matthew asked his friend Emma to go for a walk in the park. The weather was really fine so they both took their dogs to the park. Emma was so pleased to be asked by Matthew to spend time with him at the park.  Emma had a secret crush on Matthew and wondered if he felt the same but she was too scared to bring up the subject and potentially ruin their friendship. \_\_\_\_\_\_\_\_\_. This confession of feelings led to the beginning of a romantic relationship.} \newline 
 \newline
\textbf{\textit{Which resolution best fills the blank?}}
\begin{enumerate}[label=(\Alph*),topsep=0.5pt,itemsep=-0.8ex,partopsep=1ex,parsep=1ex] 
\item \textit{She never said anything about that before. It was a mystery to him.}
\item \textbf{\textit{Matthew could tell from how Emma was acting that she liked him. He turned to face her and told her how he felt.}}
\item \textit{Emma was very moved by Matthew's decision to give back the book, and agreed to accept it back. Emma let Matthew know how grateful his gesture was.}
\item \textit{She wanted to ask but didn't. He was clueless.}
\item \textit{Emma and Matthew held each others hands, kissed, and decided then that they wanted to spend the rest of their lives together.}
\end{enumerate}

%% file: 5Experiment.tex
\section{Experiments}

\subsection{Evaluation Tasks Setup}
% subtasks
For each of the two evaluation tasks, we have two context setups, varying in the amount of information provided for systems to make predictions. In Outlook Prediction, \textit{partial context} consists of only the resolution whereas \textit{full context} consists of all of the remaining stages -- seed, buildup, climax and resolution. Similarly, in Resolution Prediction, \textit{partial context} consists of only the outlook whereas \textit{full context} consists of all of the remaining stages -- seed, buildup, climax and outlook. We will refer to \textit{partial context} and \textit{full context} as \textit{context} below, indicating the corresponding narrative stages for the particular evaluation task and context setting at concern.

% confounders
We create four sets of 1250 multiple-choice questions for each context setup for each task. For each story, we pair with the correct answer four confounding choices. The first confounding choice comes from one of the four stories that share the same climax as the current story. Namely, the two stories differ only in resolution and outlook. The second confounding choice comes from one of the 24 stories that share the same buildup and the third confounding choice comes from the context of one of the 124 stories that share the same seed. The final confounding choice is from one of the 1125 stories that stem from a different seed. Below, we will refer to the confounding choices as conf-diff-seed, conf-same-seed, conf-buildup and conf-climax to notate the source of a particular confounding choice where conf-buildup means the choice comes from a story that shares the same buildup, etc. Each confounding choice is randomly picked from their respective candidate pool. The names in confounding choices that are picked from other seeds are replaced by the names of current characters. The tree structure of dataset provides a natural notion of similarity among stories, thus, the confounding choices should be of different levels of difficulties, with conf-climax being the most difficult, followed by conf-buildup, conf-same-seed and with conf-diff-seed being the easiest.

We perform 10-fold cross validation on each of the four 1250 multiple-choice question sets. Each fold consists of the 125 stories built from the same seed and we run three random restarts for each fold, taking the best predicting accuracy on the validation fold as the performance for that fold. We use the average accuracy across the 10 folds as the criterion to select hyper-parameters and report the best accuracy for each model in Tab. 8. All of our BERT-based models use BERT-base with 110M parameters, with HuggingFace's PyTorch implementation \cite{Wolf2019HuggingFacesTS}.

% add citations for word2vec, bert, huggingface
\subsection{Baseline Models}
\noindent\textbf{Average word2vec}. 
We represent both the context and each answer choice using the averaged word2vec representation of each individual word making up the sentences. We then compute the cosine similarity between each context-candidate choice pair. The answer choice that results in the highest cosine similarity value with the context is selected as the prediction. 

\clearpage
\noindent{\textbf{Pretrained Bert For Next Sentence Prediction}}. 
We use BertForNextSentencePrediction to solve the tasks where each context and each answer choice is treated as a single sequence. Each context-answer pair is assigned a single score for entailment and we evaluate the softmax of the 5 scores. The answer choice with the highest probability is chosen as the final prediction. For Resolution Prediction, we assign two scores for each answer choice: context+resolution and resolution+outlook, the average of the two scores are used to make comparisons.

\noindent{\textbf{Bert For Next Sentence Prediction with Finetuned Attention Layers}}.
For each test fold, we finetune the attention layer on the remaining 9 folds of stories and make predictions using a BertForNextSentencePrediction classifier built on top of this pretrained attention layer. 

\noindent{\textbf{Pretrained Bert For Multiple Choice}}.
We train a BerForMultipleChoice classifier similar to a classifier that solves SWAG~\cite{swag}. For Resolution Prediction full context, we concatenate other stages and outlook with a [SEP] token in between as the input context.

\noindent{\textbf{Human Evaluation}}.
We randomly select 125 MCQs for each of the evaluation tasks in full context setting and instruct MTurkers to answer these questions. Only MTurkers who are Masters and obtain a perfect score on our qualification test which consists of five sample MCQs are eligible for the annotation task. 

% Experiments Result Table
\begin{table*}[t]
\centering
\resizebox{110mm}{!}{%
\begin{tabular}{|L{3.8cm}|C{2.6cm}|C{2.6cm}|C{2.6cm}|C{2.6cm}|}
\hline
                   & \multicolumn{2}{c|}{Outlook Prediction} & \multicolumn{2}{c|}{Resolution Prediction} \\ \cline{2-5} 
                   & Resolution         & Full Story        & Outlook          & Full Story         \\ \hline
random           & 0.20             & 0.20            & 0.20           & 0.20             \\ \hline
word2vec           & 0.3184             & 0.2736            & 0.3696           & 0.3104             \\ \hline
BERTNextSent      & 0.4712             & 0.3944            & 0.4832           & 0.4464             \\ \hline
BERTNextSent(finetune) & 0.5064            & 0.4504            & 0.5096           & 0.512                \\ \hline
BertForMCQ         & 0.6088               & 0.592                & 0.6096          & 0.6352                \\ \hline
Human & N/A              &0.80                   & N/A           & 0.832              \\ \hline
\end{tabular}}
\caption{Experiment results for baseline models.}
\end{table*}
\vspace{-10pt}
% End of Experiments Result Table

%% file: 6ErrorAnalyses.tex
\section{Error Analyses}

% justify language models can't
\subsection{Distribution of Predictions}
Fig. 4 and 5 display the breakdown of predictions for BertForMultipleChoice classifier and average word2vec. Both results justify that conf-climax is the most difficult confounding choice while conf-diff-seed is the easiest with other choices of a difficulty level in between. 

\subsection{Example of Wrong Prediction}
Below is an Outlook Prediction Question that BERT-base MCQ classifier answers incorrectly. This example demonstrates that the ability to predict next sentences is not equivalent to having a social understanding. In this example, all the choices use words that are relevant to the context. The underlying relationship trajectory in each choice, however, varies significantly. A language model is unable to observe the implicit differences among choices and draw the wrong conclusion from only textual information.

\noindent\textit{Naomi's friend Noah invited her to his house. Naomi was excited to see Noah’s new place since the renovations has finally finished. Noah had cooked Naomi's favourite food. Naomi after eating a spoon of it, started to jump here and there in happiness -- Naomi loved the dessert! After eating, she expressed her gratitude to Noah for inviting her to the housewarming.} 
\vspace{0.04in} \newline 
%\textbf{\textit{Choices}}
%\vspace{0.05in} \newline
Outlook choices: \newline
\textbf{correct:} \textit{Noah felt grateful for a friend like Naomi. He asked what kind of curtains he should buy in a few weeks.} \newline
\textbf{conf-climax:} \textit{Naomi wanted to get closer to Noah and be his girlfriend.} \newline
\textbf{conf-buildup:} \textit{Naomi would pretend to enjoy the dinner but let quickly stop any of Noah's advances during the evening} \newline
\textbf{conf-same-seed:} \textit{They were both ecstatic that they each wanted a relationship with the other, and they started it that day.} \newline
\textbf{conf-diff-seed: (Bert's prediction)} \textit{Noah likes Naomi very much and wonders if this friendship can develop into something more.} \newline

\begin{figure}[H]
  \centering
  \begin{minipage}[b]{0.45\textwidth}
    \includegraphics[scale=0.5]{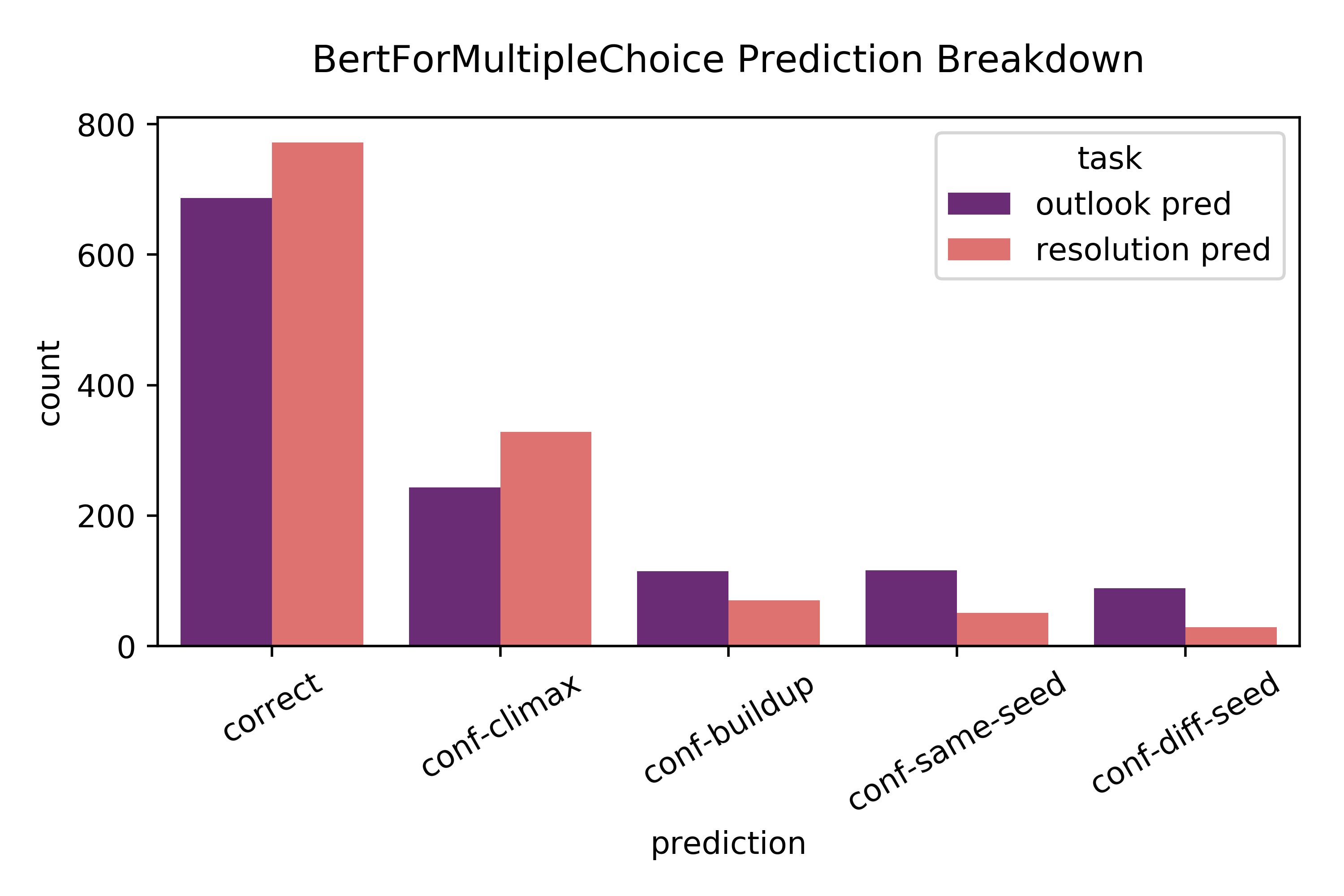}
    \caption{Bert Prediction Distribution.}
  \end{minipage}
  \hfill
  \begin{minipage}[b]{0.45\textwidth}
    \includegraphics[scale=0.5]{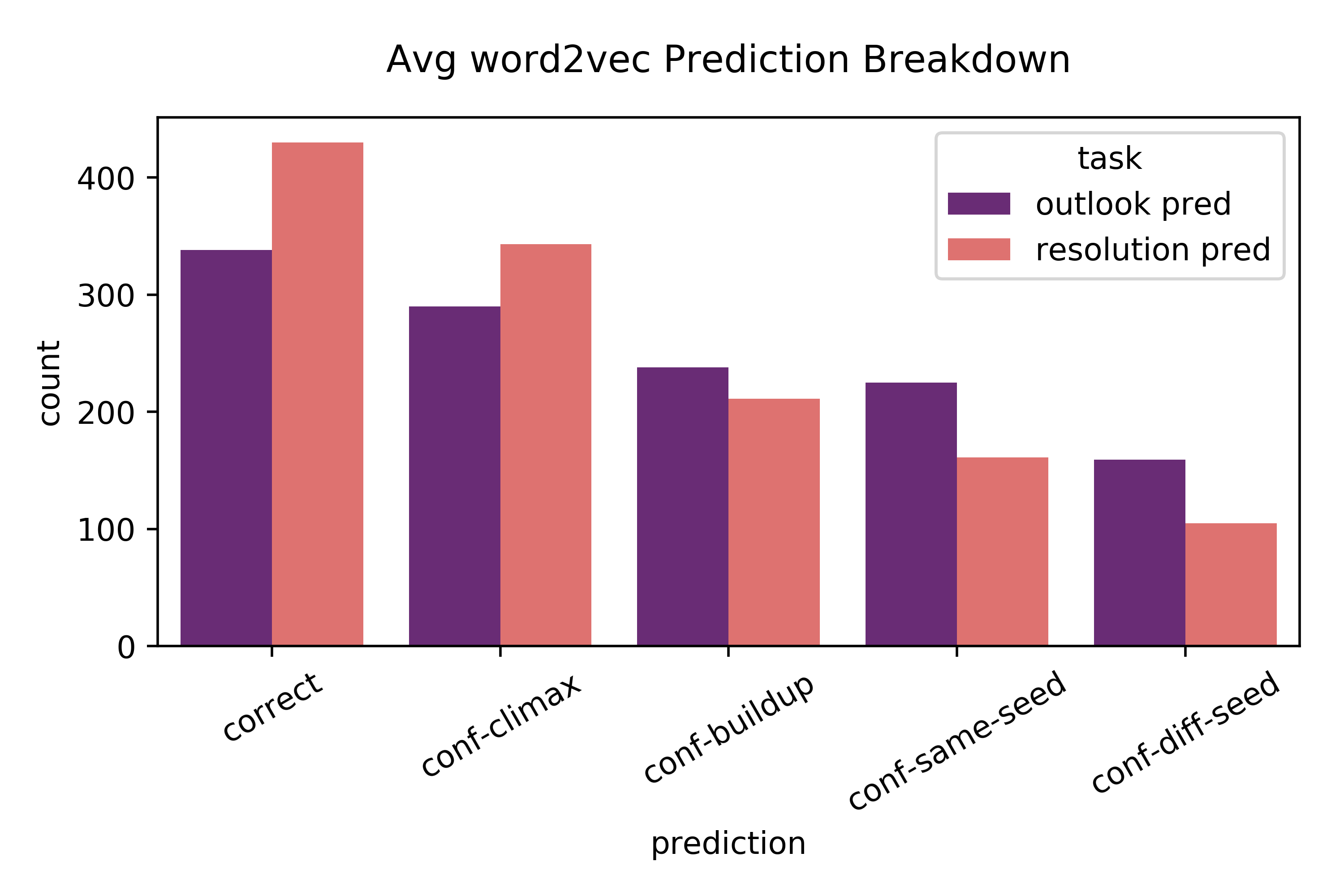}
    \caption{word2vec Prediction Distribution.}
  \end{minipage}
\end{figure}

\vspace{-10pt}
\subsection{Explanations with Social Elements}
\vspace{-2pt}
We divide the questions into two sets by whether BERT-base MCQ classifier predicts correctly. We compute the difference in VAD~\cite{VAD}, Affect Intensity~\cite{WordAffectIntensities} and SocialSent~\cite{SocialSent} scores between each correct and incorrect choice pair and compute the mean for each story, where each choice is represented by the maximum value of lexicon score for individual tokens. We then take the mean of this score across all stories and present the results in Tab. 9, 10 for VAD, Tab. 11, 12 for Affect Intensity and Tab. 13, 14 for SocialSent. For SocialSent, we display the average difference between the correct choice and each confounding choice in the corresponding column. In Tab. 9, we see that incorrectly classified questions have higher arousal difference for outlook prediction whereas for resolution prediction as shown in Tab. 10, incorrectly classified questions have higher valence and dominance difference. This shows that language modeling is unable to extract VAD information from text and use it as a hint in prediction. Similarly, in Tab. 11 and Tab. 12, the greater difference in intensity for joy and anger in outlook prediction incorrect questions and the greater difference in intensity for sadness in resolution prediction incorrect questions imply that mental state is not utilized to make predictions. In Tab. 13 and 14, the greater difference in socially contextualized sentiment score for selective choices in incorrectly answered questions is evidence that sentiment is also not taken into consideration when predictions are made. The results here further prove that language models are unable to pick up underlying social elements including VAD, Affect Intensity and mental states that set the correct answers apart from other choices.

%vad table
\begin{table}[H]
\centering
{\small
\begin{tabular}{|l|l|l|l|}
\hline
               & V      & A      & D      \\ \hline
correct   & 0.1129 & 0.1645 & 0.1325 \\ \hline
incorrect & 0.1139 & 0.1742 & 0.1274 \\ \hline
\end{tabular}
\caption{Average difference in VAD scores for outlook prediction.}
}
\end{table}
\vspace{-20pt}
\begin{table}[H]
\centering
{\small
\begin{tabular}{|l|l|l|l|}
\hline
               & V      & A      & D      \\ \hline
correct   & 0.1052 & 0.1429 & 0.1168 \\ \hline
incorrect & 0.1125 & 0.1463 & 0.1230 \\ \hline
\end{tabular}
\caption{Average difference in VAD scores for resolution prediction.}
}
\end{table}
\vspace{-20pt}

% affect intensity table
\begin{table}[H]
\centering
{\small
\begin{tabular}{|l|l|l|l|l|}
\hline
               & joy    & fear   & sadness & anger  \\ \hline
correct   & 0.3182 & 0.0627 & 0.1800  & 0.0653 \\ \hline
incorrect & 0.3223 & 0.0660 & 0.166   & 0.0726 \\ \hline
\end{tabular}
\caption{Average difference in affect intensity scores for outlook prediction.}
}
\end{table}
\vspace{-20pt}
\begin{table}[H]
\centering
{\small
\begin{tabular}{|l|l|l|l|l|}
\hline
               & joy    & fear   & sadness & anger  \\ \hline
correct   & 0.3562 & 0.1454 & 0.2810  & 0.1420 \\ \hline
incorrect & 0.3522 & 0.1492 & 0.2926  & 0.1224 \\ \hline
\end{tabular}
\caption{Average difference in affect intensity scores for resolution prediction.}
}
\end{table}
\vspace{-20pt}
\begin{table}[H]
\centering
{\small
\begin{tabular}{|l|l|l|l|l|}
\hline
conf           & climax & buildup & same-seed & diff-seed \\ \hline
correct   & 0.3599      & 0.3722       & 0.3985         & 0.3783         \\ \hline
incorrect & 0.3778      & 0.3949       & 0.3713         & 0.4026         \\ \hline
\end{tabular}
\caption{Average difference in SocialSent scores for outlook prediction.}
}
\end{table}
\vspace{-20pt}
\begin{table}[H]
{\small
\centering
\begin{tabular}{|l|l|l|l|l|}
\hline
conf        & climax & buildup & same-seed & diff-seed \\ \hline
correct   & 0.3599      & 0.3722       & 0.3985         & 0.3783         \\ \hline
incorrect & 0.3250      & 0.3400       & 0.3725         & 0.3873         \\ \hline
\end{tabular}
\caption{Average difference in SocialSent scores in resolution prediction.}
}
\end{table}

%% file: 7Summary.tex
\section{Summary}

We introduce \textit{\textbf{Social Narrative Tree}}, a corpus of 1250 social narratives that encode rich social elements that collectively drive the relationship trajectory in each story. We propose evaluation tasks Relationship Outlook Prediction MCQ and Resolution Prediction MCQ using this dataset to measure the extent that systems understand social relationships. The tree structure of the dataset provides convenient and meaningful confounding choices, making the tasks challenging for language models. Particularly, the best performance achieved by BERT-base is significantly lower than human performance, justifying that relying on textual information is insufficient in understanding social relationships. A possible future work direction is to incorporate various social elements into predicting models so as to capture the nuances of relationship trajectories in different social situations.

%% file: 0acl2020.bbl
\begin{thebibliography}{28}
\expandafter\ifx\csname natexlab\endcsname\relax\def\natexlab#1{#1}\fi

\bibitem[{Bird et~al.(2009)Bird, Klein, and Loper}]{BirdKleinLoper09}
Steven Bird, Ewan Klein, and Edward Loper. 2009.
\newblock \emph{{Natural Language Processing with Python}}.
\newblock O'Reilly Media.

\bibitem[{Bosselut et~al.(2019)Bosselut, Rashkin, Sap, Malaviya, Çelikyilmaz,
  and Choi}]{Bosselut2019COMETCT}
Antoine Bosselut, Hannah Rashkin, Maarten Sap, Chaitanya Malaviya, Asli
  Çelikyilmaz, and Yejin Choi. 2019.
\newblock Comet: Commonsense transformers for automatic knowledge graph
  construction.
\newblock \emph{ArXiv}, abs/1906.05317.

\bibitem[{Chambers and Jurafsky(2008)}]{unsupervised_learning_of_narrative}
Nathanael Chambers and Dan Jurafsky. 2008.
\newblock \href {https://www.aclweb.org/anthology/P08-1090} {Unsupervised
  learning of narrative event chains}.
\newblock In \emph{Proceedings of ACL-08: HLT}, pages 789--797, Columbus, Ohio.
  Association for Computational Linguistics.

\bibitem[{Chaturvedi et~al.(2016)Chaturvedi, Goldwasser, and
  Daume~III}]{chaturvedi2016ask}
Snigdha Chaturvedi, Dan Goldwasser, and Hal Daume~III. 2016.
\newblock Ask, and shall you receive? understanding desire fulfillment in
  natural language text.
\newblock In \emph{Thirtieth AAAI Conference on Artificial Intelligence}.

\bibitem[{Devlin et~al.(2019)Devlin, Chang, Lee, and
  Toutanova}]{devlin2019bert}
Jacob Devlin, Ming-Wei Chang, Kenton Lee, and Kristina Toutanova. 2019.
\newblock Bert: Pre-training of deep bidirectional transformers for language
  understanding.
\newblock In \emph{Proceedings of the 2019 Conference of the North American
  Chapter of the Association for Computational Linguistics: Human Language
  Technologies, Volume 1 (Long and Short Papers)}, pages 4171--4186.

\bibitem[{Duck(1994)}]{duck1994meaningful}
Steve Duck. 1994.
\newblock \emph{Meaningful relationships: Talking, sense, and relating.}
\newblock Sage Publications, Inc.

\bibitem[{Freytag(1896)}]{freytag1896freytag}
Gustav Freytag. 1896.
\newblock \emph{Freytag's technique of the drama: an exposition of dramatic
  composition and art}.
\newblock Scholarly Press.

\bibitem[{Goyal et~al.(2010)Goyal, Riloff, and
  Daum{\'e}~III}]{goyal2010automatically}
Amit Goyal, Ellen Riloff, and Hal Daum{\'e}~III. 2010.
\newblock Automatically producing plot unit representations for narrative text.
\newblock In \emph{Proceedings of the 2010 Conference on Empirical Methods in
  Natural Language Processing}, pages 77--86.

\bibitem[{Granroth-Wilding and Clark(2016)}]{WhatHappensNext}
Mark Granroth-Wilding and Stephen Clark. 2016.
\newblock What happens next? event prediction using a compositional neural
  network model.
\newblock In \emph{AAAI}.

\bibitem[{Hamilton et~al.(2016)Hamilton, Clark, Leskovec, and
  Jurafsky}]{SocialSent}
William~L. Hamilton, Kevin Clark, Jure Leskovec, and Dan Jurafsky. 2016.
\newblock \href {http://arxiv.org/abs/1606.02820} {Inducing domain-specific
  sentiment lexicons from unlabeled corpora}.
\newblock \emph{CoRR}, abs/1606.02820.

\bibitem[{Han et~al.(2019)Han, Choi, and Tan}]{han2019permanent}
Xiaochuang Han, Eunsol Choi, and Chenhao Tan. 2019.
\newblock \href {http://arxiv.org/abs/1904.08950} {No permanent friends or
  enemies: Tracking relationships between nations from news}.

\bibitem[{Iyyer et~al.(2016)Iyyer, Guha, Chaturvedi, Boyd-Graber, and
  Daum{\'e}~III}]{iyyer2016feuding}
Mohit Iyyer, Anupam Guha, Snigdha Chaturvedi, Jordan Boyd-Graber, and Hal
  Daum{\'e}~III. 2016.
\newblock Feuding families and former friends: Unsupervised learning for
  dynamic fictional relationships.
\newblock In \emph{Proceedings of the 2016 Conference of the North American
  Chapter of the Association for Computational Linguistics: Human Language
  Technologies}, pages 1534--1544.

\bibitem[{Labov(1997)}]{Labov1997SomeFS}
William Labov. 1997.
\newblock Some further steps in narrative analysis.

\bibitem[{Loria et~al.(2014)Loria, Keen, Honnibal, Yankovsky, Karesh, Dempsey
  et~al.}]{loria2014textblob}
Steven Loria, P~Keen, M~Honnibal, R~Yankovsky, D~Karesh, E~Dempsey, et~al.
  2014.
\newblock Textblob: simplified text processing.
\newblock \emph{Secondary TextBlob: simplified text processing}, 3.

\bibitem[{Mohammad(2018)}]{VAD}
Saif Mohammad. 2018.
\newblock \href {https://doi.org/10.18653/v1/P18-1017} {Obtaining reliable
  human ratings of valence, arousal, and dominance for 20,000 {E}nglish words}.
\newblock In \emph{Proceedings of the 56th Annual Meeting of the Association
  for Computational Linguistics (Volume 1: Long Papers)}, pages 174--184,
  Melbourne, Australia. Association for Computational Linguistics.

\bibitem[{Mohammad(2017)}]{WordAffectIntensities}
Saif~M. Mohammad. 2017.
\newblock \href {http://arxiv.org/abs/1704.08798} {Word affect intensities}.
\newblock \emph{CoRR}, abs/1704.08798.

\bibitem[{Mostafazadeh et~al.(2016)Mostafazadeh, Chambers, He, Parikh, Batra,
  Vanderwende, Kohli, and Allen}]{rocstories}
Nasrin Mostafazadeh, Nathanael Chambers, Xiaodong He, Devi Parikh, Dhruv Batra,
  Lucy Vanderwende, Pushmeet Kohli, and James Allen. 2016.
\newblock \href {https://doi.org/10.18653/v1/N16-1098} {A corpus and cloze
  evaluation for deeper understanding of commonsense stories}.
\newblock In \emph{Proceedings of the 2016 Conference of the North {A}merican
  Chapter of the Association for Computational Linguistics: Human Language
  Technologies}, pages 839--849, San Diego, California. Association for
  Computational Linguistics.

\bibitem[{Prince(1973)}]{prince1973grammar}
G.~Prince. 1973.
\newblock \href {https://books.google.com/books?id=UDY6pRQSFAYC} {\emph{A
  Grammar of Stories: An Introduction}}.
\newblock De proprietatibus litterarum. Mouton.

\bibitem[{Rahimtoroghi et~al.(2017)Rahimtoroghi, Wu, Wang, Anand, and
  Walker}]{rahimtoroghi2017modelling}
Elahe Rahimtoroghi, Jiaqi Wu, Ruimin Wang, Pranav Anand, and Marilyn Walker.
  2017.
\newblock Modelling protagonist goals and desires in first-person narrative.
\newblock In \emph{Proceedings of the 18th Annual SIGdial Meeting on Discourse
  and Dialogue}, pages 360--369.

\bibitem[{Rashkin et~al.(2018{\natexlab{a}})Rashkin, Bosselut, Sap, Knight, and
  Choi}]{RocstoryPsychology}
Hannah Rashkin, Antoine Bosselut, Maarten Sap, Kevin Knight, and Yejin Choi.
  2018{\natexlab{a}}.
\newblock \href {http://arxiv.org/abs/1805.06533} {Modeling naive psychology of
  characters in simple commonsense stories}.
\newblock \emph{CoRR}, abs/1805.06533.

\bibitem[{Rashkin et~al.(2018{\natexlab{b}})Rashkin, Sap, Allaway, Smith, and
  Choi}]{rashkin2018event2mind}
Hannah Rashkin, Maarten Sap, Emily Allaway, Noah~A. Smith, and Yejin Choi.
  2018{\natexlab{b}}.
\newblock Event2mind: Commonsense inference on events, intents, and reactions.
\newblock In \emph{ACL}.

\bibitem[{Rubin et~al.(1995)Rubin, Booth-LaForce, Rose-Krasnor, and
  Mills}]{rubin1995competence}
Kenneth Rubin, Cathryn Booth-LaForce, Linda Rose-Krasnor, and Rosemary Mills.
  1995.
\newblock Social relationships and social skills: A conceptual and empirical
  analysis.

\bibitem[{Sap et~al.(2018)Sap, LeBras, Allaway, Bhagavatula, Lourie, Rashkin,
  Roof, Smith, and Choi}]{Atomic}
Maarten Sap, Ronan LeBras, Emily Allaway, Chandra Bhagavatula, Nicholas Lourie,
  Hannah Rashkin, Brendan Roof, Noah~A. Smith, and Yejin Choi. 2018.
\newblock \href {http://arxiv.org/abs/1811.00146} {{ATOMIC:} an atlas of
  machine commonsense for if-then reasoning}.
\newblock \emph{CoRR}, abs/1811.00146.

\bibitem[{Sap et~al.(2019)Sap, Rashkin, Chen, LeBras, and
  Choi}]{sap2019socialIQa}
Maarten Sap, Hannah Rashkin, Derek Chen, Ronan LeBras, and Yejin Choi. 2019.
\newblock Social iqa: Commonsense reasoning about social interactions.
\newblock In \emph{EMNLP}.

\bibitem[{Schank and Abelson(1977)}]{ScriptSchankAbelson1977}
R.C. Schank and R.~Abelson. 1977.
\newblock \emph{Scripts, Plans, Goals, and Understanding}.
\newblock Hillsdale, NJ: Earlbaum Assoc.

\bibitem[{Wolf et~al.(2019)Wolf, Debut, Sanh, Chaumond, Delangue, Moi, Cistac,
  Rault, Louf, Funtowicz, and Brew}]{Wolf2019HuggingFacesTS}
Thomas Wolf, Lysandre Debut, Victor Sanh, Julien Chaumond, Clement Delangue,
  Anthony Moi, Pierric Cistac, Tim Rault, R'emi Louf, Morgan Funtowicz, and
  Jamie Brew. 2019.
\newblock Huggingface's transformers: State-of-the-art natural language
  processing.
\newblock \emph{ArXiv}, abs/1910.03771.

\bibitem[{Zellers et~al.(2018)Zellers, Bisk, Schwartz, and Choi}]{swag}
Rowan Zellers, Yonatan Bisk, Roy Schwartz, and Yejin Choi. 2018.
\newblock \href {http://arxiv.org/abs/1808.05326} {{SWAG:} {A} large-scale
  adversarial dataset for grounded commonsense inference}.
\newblock \emph{CoRR}, abs/1808.05326.

\bibitem[{Zellers et~al.(2019)Zellers, Holtzman, Bisk, Farhadi, and
  Choi}]{HellaSwag}
Rowan Zellers, Ari Holtzman, Yonatan Bisk, Ali Farhadi, and Yejin Choi. 2019.
\newblock \href {http://arxiv.org/abs/1905.07830} {Hellaswag: Can a machine
  really finish your sentence?}
\newblock \emph{CoRR}, abs/1905.07830.

\end{thebibliography}
